\documentclass{article}

\usepackage{PRIMEarxiv}

\usepackage{amsmath,amsfonts}
\usepackage{algorithmic}
\usepackage{algorithm}
\usepackage{array}
\usepackage[caption=false,font=normalsize,labelfont=sf,textfont=sf]{subfig}
\usepackage{textcomp}
\usepackage{stfloats}
\usepackage{url}
\usepackage{verbatim}
\usepackage{graphicx}
\usepackage{cite}

\usepackage{colortbl}
\usepackage{xcolor}
\usepackage{color}
\usepackage{multirow}

\title{A Knowledge-guided Adversarial Defense for Resisting Malicious Visual Manipulation
}

\author{
  \;\;Dawei Zhou\;\; \\
  \;\;Xidian University\;\; \\
  \And
\;\;Zhigang Su\;\; \\
  \;\;Xidian University\;\;\\
  \And
 \;\;Decheng Liu\;\;\\
  \;\;Xidian University\;\;\\
   \And
   \And
 Tongliang Liu\\
  The University of Sydney\\
    \And
 Nannan Wang*\\
  Xidian University\\
    \And
 Xinbo Gao\\
  Xidian University\\
}

\begin{document}
\maketitle

\begin{abstract}
Malicious applications of visual manipulation have raised serious threats to the security and reputation of users in many fields. To alleviate these issues, adversarial noise-based defenses have been enthusiastically studied in recent years. However, ``data-only" methods tend to distort fake samples in the low-level feature space rather than the high-level semantic space, leading to limitations in resisting malicious manipulation. Frontier research has shown that integrating knowledge in deep learning can produce reliable and generalizable solutions. Inspired by these, we propose a \textit{knowledge-guided adversarial defense (KGAD) to actively force malicious manipulation models to output semantically confusing samples}. Specifically, in the process of generating adversarial noise, we focus on constructing significant semantic confusions at the domain-specific knowledge level, and exploit a metric closely related to visual perception to replace the general pixel-wise metrics. The generated adversarial noise can actively interfere with the malicious manipulation model by triggering knowledge-guided and perception-related disruptions in the fake samples. To validate the effectiveness of the proposed method, we conduct qualitative and quantitative experiments on human perception and visual quality assessment. The results on two different tasks both show that our defense provides better protection compared to state-of-the-art methods and achieves great generalizability.
\end{abstract}


\section{Introduction}
\label{sec1}

With the rapid development of deep generative techniques (\textit{e.g.} generative adversarial networks \cite{goodfellow2020generative} and variational autoencoders \cite{kingma2013auto}), visual manipulation has achieved impressive achievements, creating considerable cultural and economic value. A variety of visual manipulation methods have been proposed, especially in the fields of face manipulation \cite{he2019attgan,tang2019attention,li2021image} and style manipulation \cite{chu2017cyclegan,choi2018stargan,zhu2022all}. 
As the generated results had increasingly realistic quality and even fooled human eyes, manipulation techniques are easily misused for malicious purposes, such as invading personal privacy \cite{korshunov2018deepfakes} and misleading public opinion \cite{tolosana2020deepfakes}. In detail, malicious users can edit portrait appearances, forge identities or alter important information without permission. These malicious applications raise serious security and reputation threats in society.

\begin{figure}[t]
\begin{center}
   \includegraphics[width=0.6\linewidth]{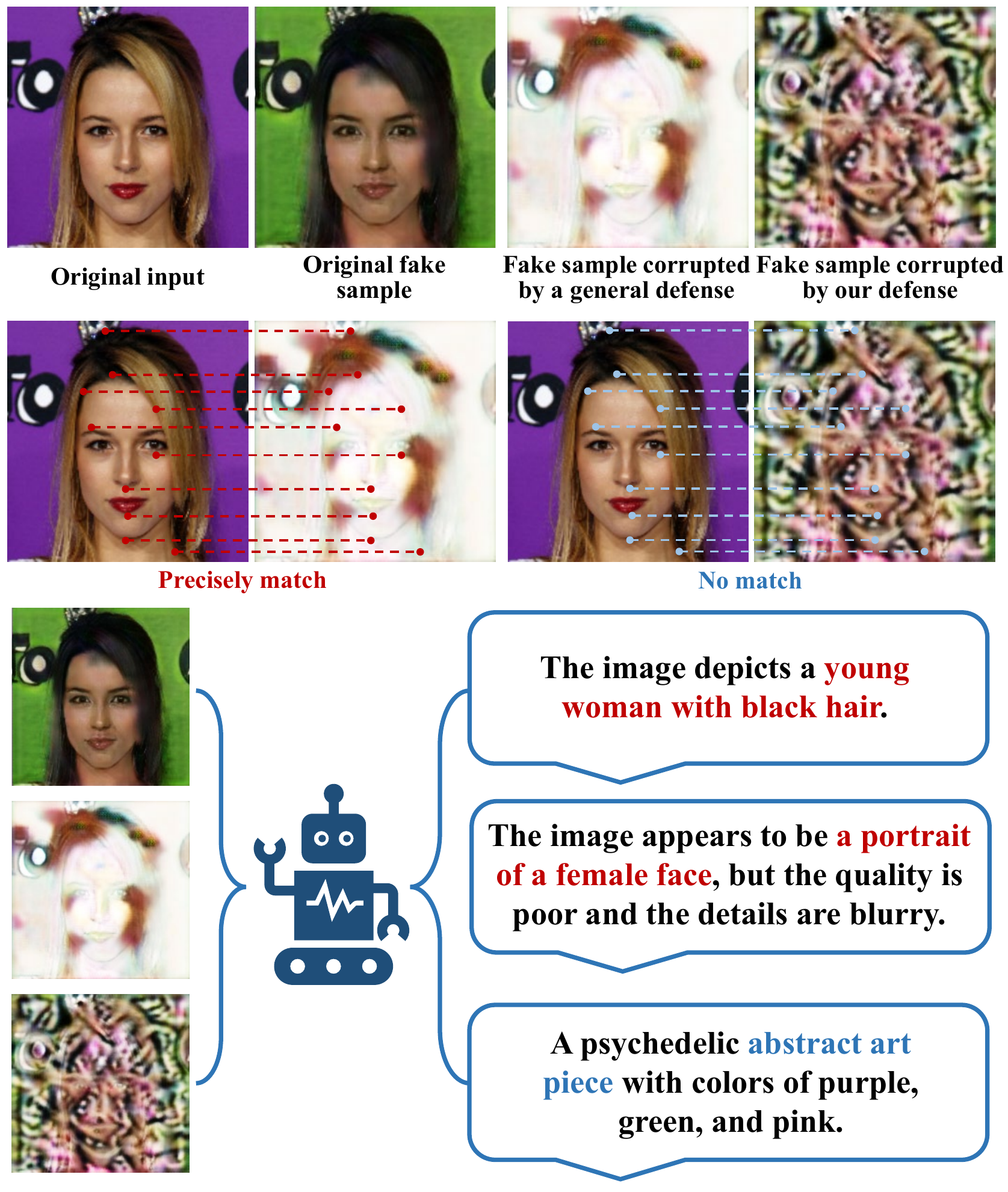}
\end{center}
   \caption{Distorted fake samples corrupted by adversarial noise-based defenses. The top third and fourth figures are corrupted by the general method and our proposed method, respectively. The disruptions in the former are mainly clustered in local color textures while the overall structure is still clear. Conversely, the disruptions in the latter significantly perturb the semantic information (\textit{e.g.}, face structure), which causes more confusion from the perspective of human vision. The face in the fake sample corrupted by the general defense precisely matches the face in the original input (see the middle figures), leading that identity privacy is still used for malicious actions, but our method mitigates this issue. Moreover, we use an intelligent model to understand the content of an image. According to the statements of the model for the three samples, it can be seen that the general defense leaves out critical information, while our method performs a more sufficient obfuscation.}
\label{fig1}
\end{figure}

To mitigate the above issues, defensive measures against malicious visual manipulation have been widely studied. Deepfake detection is a major strategy which is able to achieve high accuracy in discriminating fake samples \cite{lyu2014exposing,rossler2019faceforensics++,qian2020thinking,dang2020detection,mirsky2021creation}. Unfortunately, this ex-post passive approach cannot essentially eliminate the harm of malicious visual manipulation because it is hard to prevent the normal generation of fake samples. How to actively combat the threat of deepfake is an important but not yet sufficiently explored problem. 
Recently, a type of adversarial noise-based defenses \cite{ruiz2020disrupting,yeh2020disrupting,wang2022anti,aneja2022tafim,huang2021initiative} shift targets from the data to the manipulation procedure itself. They embed imperceptible adversarial noise in input samples to interfere with malicious manipulation models, causing them to produce distorted outputs, making the forgery fail. However, existing methods are usually in a ``data-only" manner and important knowledge in visual manipulation has not been deeply considered, which may lead to limitations in defense effect.

\begin{figure*}[t]
\begin{center}
   \includegraphics[width=0.8\linewidth]{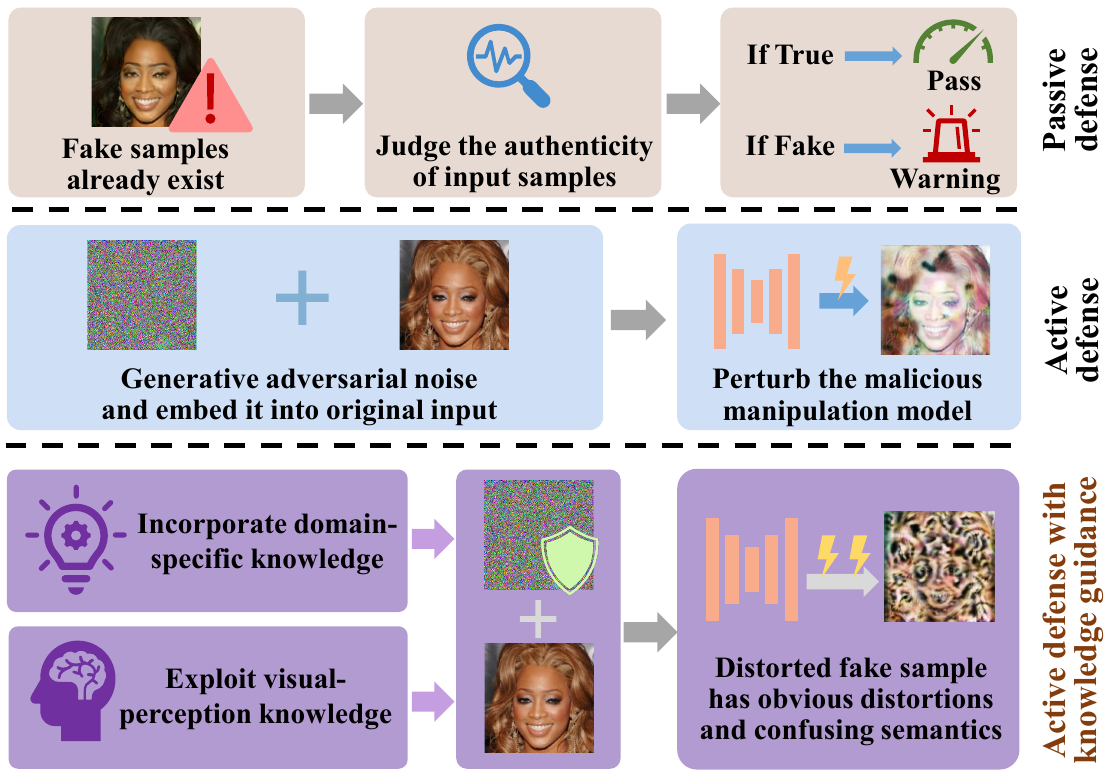}
\end{center}
   \caption{The purpose of the proposed method. Unlike the detection-based passive defense (top), our method (bottom) focuses on embedding human-imperceptible noise into input samples to perturb malicious manipulation models. Moreover, different with general active defenses (middle), the adversarial noise of our method is constructed under the guidance of domain-specific and visual-perception knowledge to make the distorted fake samples have obvious anomalies and confusing semantics, so that critical information (\textit{e.g.}, the identity privacy) is more thoroughly obfuscated.}
\label{fig2}
\end{figure*}

Specifically, researches on knowledge discovery \cite{ graves2012long,lazer2014parable,marcus2014eight,karpatne2022knowledge} have shown methods that rely solely on data are not closely tied to underlying scientific theories. They tend to exploit low-level features rather than high-level semantics, and thus are susceptible to producing solutions that are inconsistent with existing knowledge. Existing adversarial noise-based defenses mainly focus on maximizing disruptions in the low-level space, such as pixel-wise mean square error. On the one hand, such defenses \textit{lack guidance from domain-specific knowledge}. Domain-specific knowledge is defined as declarative, procedural, or conditional knowledge related to a particular field \cite{alexander1988interaction} and can lead to action permitting specified task completion \cite{tricot2014domain}. It can facilitate the capture of key information and make it a lot easier to explain the results. The lack of such knowledge prevents the defense from perturbing the core content (\textit{e.g.}, face structure) of fake samples, resulting a failure to constitute significant distortions from the perspective of human cognition (see Figure~\ref{fig1}). On the other hand, \textit{the knowledge related to visual perception is not sufficiently incorporated}, so that the resulting anomalies are usually clustered in local color textures while the overall structure under the anomalous textures is still relatively normal. These limitations restrict the effectiveness of defenses in protecting important and private visual data.

In addition, frontier works have indicated that integrating scientific knowledge in deep learning can help produce reliable and generalizable solutions \cite{karpatne2017theory,karpatne2022knowledge}. Motivated by this, we propose a \textit{Knowledge-Guided Adversarial Defense (KGAD) against malicious visual manipulation} (see Figure~\ref{fig2}). The proposed method is committed to design protective adversarial noise for actively forcing malicious manipulation models to output significantly anomalous and semantically confusing samples, so that critical information (\textit{e.g.}, the identity privacy) is more thoroughly obfuscated. Obviously distorted fake samples will not be well utilized for malicious actions such as invading the privacy and misleading the public. 

In detail, in the process of generating adversarial noise, we focus on constructing significant confusions in the semantic space of domain-specific knowledge. For example, for the face (or style) manipulation, we calculate the distance between the face keypoints (or content features) of the original fake sample and the distorted fake sample, and then utilize it as a semantic guidance. Moreover, we replace the general low-level pixel-wise mean square error with a metric that is closely related to visual-perception knowledge, such as Structural Similarity Index Measure (SSIM) \cite{wang2004image}. Considering that \textit{perception knowledge is applicable to different vision tasks} and \textit{domain-specific knowledge is commonly used for feature construction and selection in corresponding deep learning models} \cite{karpatne2022knowledge,congdon2007bayesian}, we believe that the generated adversarial noise can more effectively and generalizably interfere with malicious manipulation models by maximizing domain-specific and perception-related disruptions. 

To verify the effectiveness of the proposed method, experiments on two types of vision tasks (\textit{i.e.}, face manipulation and style manipulation) are conducted. We perform qualitative and quantitative evaluations from the human perception and visual quality assessment perspectives, respectively. 

The main contributions of this work are as follows:
\begin{itemize}
\item{We introduce the knowledge guidance to actively defend against the malicious visual manipulation, which provides a new perspective for adversarial noise-based defense. We hope this mechanism can alleviate the limitations of ``data-only" defenses and inspire more great works for this worthwhile field.}

\item{We propose a knowledge-guided adversarial defense, which aims to actively force malicious manipulation models to output significantly anomalous and semantically confusing samples, so that critical information is more thoroughly obfuscated. We perform disruption maximization at the level of domain-specific and visual-perception knowledge to obtain protective adversarial noise.}

\item{Extensive experiments are conducted to demonstrate the effectiveness of the proposed defense. Qualitative and quantitative evaluations show that the adversarial noise generated by our method exhibits better defensive capabilities and generalization for input samples against malicious manipulation models.}
\end{itemize}

\section{Related works}
\label{sec2}
\noindent\textbf{Visual manipulation.} 
The tremendous success of deep generative models has enabled visual manipulation to synthesize detailed and realistic images \cite{karras2018progressive,brocklarge,karras2020analyzing,karras2021alias,liang2023pmsgan}. However, visual manipulation techniques might be used by malicious actors for unethical behaviors. Representatively, face manipulation \cite{he2019attgan,tang2019attention,li2021image} can modify face attributes or even synthesize new faces to falsify identity, and style manipulation \cite{chu2017cyclegan,choi2018stargan,zhu2022all} can tamper with visual information in important or sensitive data. These issues raise serious potential threats such as invading personal privacy and inciting public opinion, posing serious challenges to maintaining the security and reputation of society. Therefore, it is meaningful and urgent to find effective defenses against malicious visual manipulation. In this work, we select five classic deep learning models as the manipulation models to participate in the evaluation of defenses. StarGAN \cite{choi2018stargan} proposes a scalable approach to perform image-to-image translation across different domains and its generated samples obtain great visual quality. 
AGGAN \cite{tang2019attention} utilizes a built-in attention mechanism to introduce attention masks for crafting target images with high quality. HiSD \cite{li2021image} is an advanced image-to-image translation model, it has impressive disentanglement and controllable diversity. These models adopt different architectures and losses and are thus able to convincingly reflect the effectiveness and generalization of the defenses.

\noindent\textbf{Malicious manipulation defense.}
The rapidly increasing incidents of malicious visual manipulation and their serious hazards have prompted a growing need for defensive measures. Most of existing defenses are deepfake detection-based methods \cite{lyu2014exposing,rossler2019faceforensics++,qian2020thinking,dang2020detection,mirsky2021creation}, which belong to the post-hoc passive defense mechanisms. Some traditional fake detection techniques exploit hand-crafted features (such as gradients or compression artifacts) to find inconsistent visual information \cite{lyu2014exposing,agarwal2017photo}. However, as the fidelity of visual manipulation increases substantially, their accuracy has declined. Subsequently, learning-based detection methods \cite{agarwal2019protecting,rossler2019faceforensics++,aneja2020generalized,qian2020thinking,mirsky2021creation,cozzolino2021id} are proposed, which are able to spot fake samples with high confidences. 

Recently, with the development of adversarial learning, some researchers have turned their attention from fake data to manipulation models themselves and propose adversarial noise-based defense mechanisms \cite{ruiz2020disrupting,yeh2020disrupting,aneja2022tafim,huang2021initiative}. This type of defense achieves active protection of data by generating adversarial noise \cite{goodfellow2014explaining,madry2017towards} to distort the output of the malicious manipulation model. Ruiz \textit{et al.}~\cite{ruiz2020disrupting} defends against malicious manipulation by maximizing the Mean Square Error between the fake samples corresponding to the original input and the protected input. Huang \textit{et al.} \cite{huang2021initiative} designs a two-stage training framework to construct an initiative defense. Aneja \textit{et al.} \cite{aneja2022tafim} trains a generative model to obtain the adversarial noise that skews the output of the manipulation model toward one color.
However, these methods belong to the ``data-only" defense manner and the important knowledge of the specific domain and visual perception has not been well considered. Differently, our method incorporated the guidance of domain-specific knowledge and visual-perception knowledge into the generation of adversarial noise, which is expected to further enhance the defensive effectiveness.

\section{Methodology}
\label{sec3}
In this section, we first provide the preliminaries on the fundamental of adversarial noise-based defense and then illustrate the proposed method.

\subsection{Adversarial noise-based defense}
\label{sec3.1}
Adversarial-noise based defense is a recently proposed new mechanism to combat the malicious visual manipulation. This mechanism generates human-imperceptible but adversarial perturbations (\textit{i.e.}, adversarial noise) and adds them to the original input sample to interfere with the malicious visual manipulation model. The disruption on the output can be regarded as the result of an adversarial attack on the manipulation model and males the manipulation model lose the ability to craft realistic fake samples. That is, the fake sample has obvious anomalies and is highly unrealistic in human vision. 

Formulaically, let $x \in \mathbb{R}^{H \times W \times C}$ denote the original input sample with the height $H$, width $W$ and channel $C$. Let $\delta \in \mathbb{R}^{H \times W \times C}$ denotes the adversarial noise which is usually constrained to a perturbation budget for human-imperceptibility, \textit{i.e.}, $\|\delta\| \leq \epsilon$ where $\| \cdot \|$ denotes the norm constraint (\textit{e.g.}, $L_{\infty}$-norm: $\| \cdot \|_{\infty}$). The adversarial noise $\delta$ is embedded in the original input sample $x$ and produces an adversarial protected sample $x^{\prime} \in \mathbb{R}^{H \times W \times 3}$ where $x^{\prime} = x + \delta$. We denote the malicious manipulation function by $g:\mathbb{R}^{H \times W \times 3} \rightarrow \mathbb{R}^{H \times W \times 3}$. The manipulation function $g$ can be parameterized via a deep neural network $G_\theta(\cdot)$ where $\theta$ is its model parameters. Given an original input sample $x$ and an adversarial protected sample $x^{\prime}$, the manipulation model $G_\theta$ outputs the original fake sample $y \in \mathbb{R}^{H \times W \times C}$ and the distorted fake sample $y^{\prime} \in \mathbb{R}^{H \times W \times C}$ (\textit{i.e.}, $y=G_\theta(x), y^{\prime}=G_\theta(x^{\prime})$). The objective function of the $\delta$ can be formulated as:
\begin{equation}
\label{eq1}
\max _\delta \space dis(G_\theta(x), G_\theta(x+\delta)) \text {, s.t. } \|\delta\| \leq \epsilon,
\end{equation}
where $dis(\cdot,\cdot)$ is a distance metric, such as $L_p$-norm. The optimization of $\delta$ can be effectively executed by using Fast Gradient Sign Method (FGSM) \cite{goodfellow2014explaining}, Projected Gradient Descent (PGD) \cite{madry2017towards} or other attack strategies. Among them, as the strongest first-order attack, PGD is widely used in adversarial-noise based defenses.

\subsection{Knowledge-guided adversarial defense}
\label{sec3.2}
For the Equation.~\ref{eq1}, adopting $L_p$-norm as the distance metric is a direct and simple choice. Most of adversarial noise-based defenses directly use the $L_p$-norm between the original fake sample $y$ and the distorted fake sample $y^{\prime}$ as the main criterion to judge whether the protection is successful (\textit{i.e.}, whether the adversarial noise effectively interferes with the malicious manipulation model to distort its output). Although this mechanism is able to cause anomalies to fake samples, it still had some limitations in terms of defense effectiveness.

\begin{figure}[t]
\begin{center}
   \includegraphics[width=0.5\linewidth]{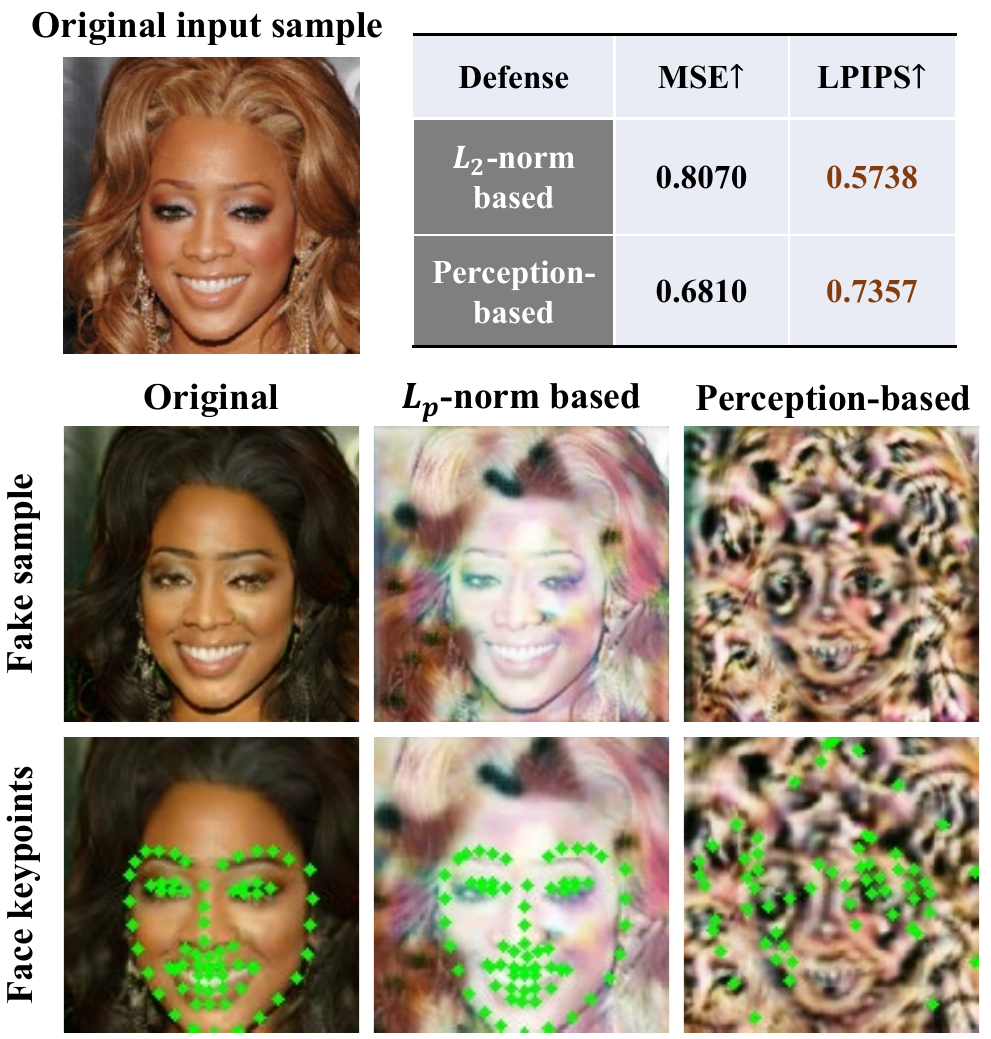}
\end{center}
   \caption{The limitations of general adversarial noise-based defenses. Although the distorted fake samples have anomalies compared with original fake samples, the face appearance is still clearly visible and the keypoints is normally detected. Furthermore, although maximizing $L_2$-norm can lead to a larger value on pixel-wise MSE, it does not maintain this advantage on the LPIPS indicator which is \textit{more} consistent with human vision.}
\label{fig3}
\end{figure}

\textit{On the one hand}, domain-specific knowledge is not sufficiently integrated into the process of generating adversarial noise, which leads to defenses failing to deeply focus on high-level semantic information and effectively construct confusions in the semantic space. For example, as shown in Figure~\ref{fig3}, although the distorted fake sample has anomalies compared with the original fake sample, the face appearance is still clearly visible and the keypoints can be normally detected. This indicates that the defense does not essentially achieve the semantic destruction to the malicious manipulation model from the perspective of human cognition. \textit{On the other hand}, such mechanism usually ignores the guidance from the visual-perception knowledge, allowing the variations brought by adversarial noise are mainly concentrated on local textures or present in the form of color flipping. The overall structure under anomalous textures is still relatively normal. 

In addition, previous studies \cite{wang2004image,hore2010image,zhang2011fsim,sharif2018suitability,zhang2018unreasonable,sara2019image,hameed2021perceptually} have shown that the $L_p$-norm (with $p \in {1, 2, \infty}$) metric lacks enough suitability for perception-consistent visual quality assessment. It does not capture the perceptual quality of samples \cite{hameed2021perceptually}, and having a small $L_p$ distance is both unnecessary and insufficient for perceptual similarity \cite{sharif2018suitability}. As shown in Figure~\ref{fig3}, the Mean Square Error (MSE) between the distorted fake sample and original fake sample is larger under the $L_2$-norm based constraint than under a perception-based metric (\textit{e.g.}, SSIM). 
However, the latter causes more significant semantic confusions from the perspective of human eyes and has better performances in an indicator that is more consistent with human perception, \textit{e.g.}, the Learned Perceptual Image Patch Similarity (LPIPS) \cite{zhang2018unreasonable}.

\begin{figure*}[t]
\begin{center}
   \includegraphics[width=0.95\linewidth]{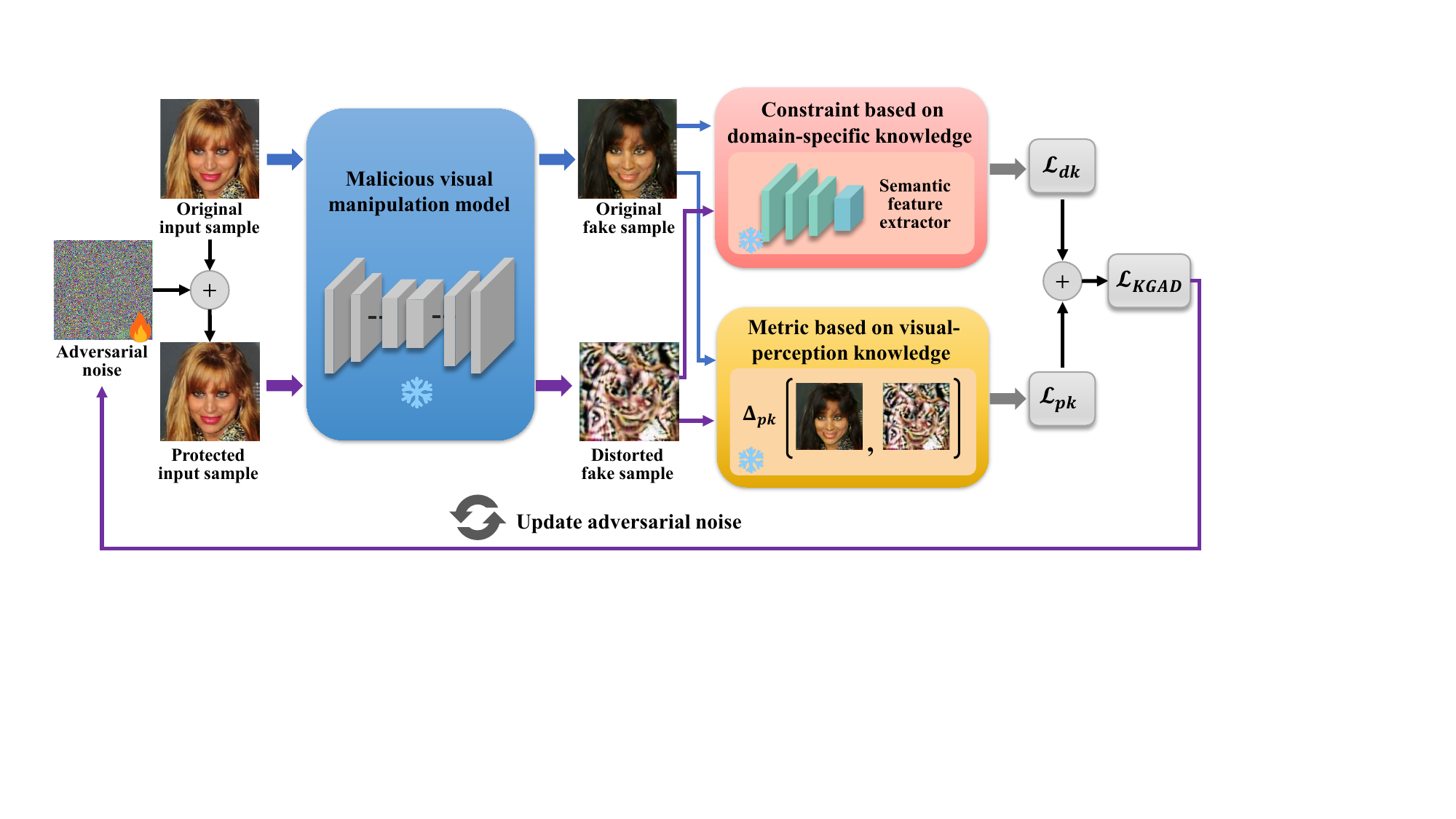}
\end{center}
   \caption{The schematic diagram of the proposed knowledge-guided adversarial defense method. We leverage the knowledge guidance to assist in the generation of protective adversarial noise, and then add the noise to the original input sample to interfere with the malicious manipulation model, making it produce obvious distortions and confusing semantics. The proposed method consists of two main components: a constraint based on the domain-specific knowledge and a metric based on the visual-perception knowledge. The former works on destroying important semantics associated with the specific task (\textit{e.g.}, the face structure for the face manipulation task) in the fake samples, and the latter focuses on disrupting visual perception-related features. Our method strives to achieve the above goals by jointly minimizing the domain-specific loss $\mathcal{L}_{dk}$ and the visual-perception loss $\mathcal{L}_{pk}$ to iteratively update the adversarial noise.}
\label{fig4}
\end{figure*}

The above observations suggest that relying solely on metrics on the data pixels themselves may make the defenses tend to focus on low-level features at the expense of ignoring high-level semantics, which is consistent with the researches on knowledge discovery \cite{karpatne2017theory,karpatne2022knowledge}. The protective noise thus has limited effect in interfering with malicious manipulation models. Based on these, in this work, we propose a \textit{Knowledge-Guided Adversarial Defense (KGAD)} to actively force malicious manipulation models to output fake samples which are significantly anomalous and semantically confusing. The proposed method consists of two main components: a constraint based on the domain-specific knowledge and a metric based on the visual-perception knowledge. The schematic diagram of the proposed method is shown in Figure~\ref{fig4}

\noindent\textbf{The constraint based on domain-specific knowledge.} 
In order to enable the generated adversarial noise to leverage high-level semantics to interference with malicious manipulation models, we construct a new loss function from the perspective of domain-specific knowledge:
\begin{equation}
\label{eq2}
\mathcal{L}_{dk} = - \ell_d(\mathcal{K}_d(G_\theta(x)), \mathcal{K}_d(G_\theta(x+\delta))),
\end{equation}
where $\mathcal{K}_d(\cdot)$ denotes the extracted semantic features of domain-specific knowledge and $\ell_d(\cdot,\cdot)$ denotes standard distance metric (\textit{e.g.}, MSE). The loss function from the domain-specific knowledge $\mathcal{L}_{dk}$ can help extract corresponding semantic features for the tasks in different domains. For example, $\mathcal{K}_d(\cdot)$ is the keypoint semantic of the face for the malicious face manipulation, and it is the content semantics of the object for the malicious style manipulation. The former can be obtained by a keypoint detection model such as an improved MobileFaceNet \cite{PFL} and the latter can be obtained by a content extraction model such as VGGNet-16 \cite{simonyan2014very,Gatys_2016_CVPR}.


\noindent\textbf{The metric based on visual-perception knowledge.} 
Besides the constraint of domain-specific knowledge, we exploit a metric related to visual-perception knowledge to replace the $L_p$-norm in Equation.~\ref{eq1}. The loss function from the visual-perception knowledge is formulated as:
\begin{equation}
\label{eq3}
\mathcal{L}_{pk} = - \Delta_{pk}(G_\theta(x), G_\theta(x+\delta)),
\end{equation}
where $\Delta_{pk}(\cdot,\cdot)$ denotes a metric that is more consistent with human vision than $L_2$-norm, such as the SSIM Dissimilarity (SSIMD) or LPIPS. In this work, we exploit SSIMD as $\Delta_{pk}$ and take the better LPIPS as an evaluation indicator for fairly evaluating different defense methods.

\noindent\textbf{Optimization algorithm.} 
On the basis of the above Equations~\ref{eq2} and ~\ref{eq3}, the overall loss function of the proposed method is formulated as
\begin{equation}
\label{eq4}
\mathcal{L}_{KGAD} = \mathcal{L}_{pk} + \lambda \cdot \mathcal{L}_{dk},
\end{equation} 
where $\lambda$ is the trade-off hyperparameter. 
The optimization algorithm is shown in Algorithm.~\ref{alg1}. 
In detail, for each input $x_i$ sampled from the dataset, we first obtain an initial protective adversarial noise $\delta_i^{0}$. Then, in $t$-th iteration, we embed $\delta_i^{t}$ into the input sample $x_i$ and obtain the adversarial sample $x_i^{\prime_t}=x_i+\delta_i^{t}$. Next, we forward pass $x_i$ and $x_i^{\prime_t}$ into the malicious manipulation model $G_\delta$ and get corresponding original fake sample $y_i=G_\delta(x_i)$ and distorted fake sample $y_i^{\prime_t}=G_\delta(x_i+\delta_i^{t})$. After that, we compute the loss function based on the domain-specific knowledge $\mathcal{L}_{dk}$ via Equation.~\ref{eq2}, the loss function based on the visual-perception knowledge $\mathcal{L}_{pk}$ via Equation.~\ref{eq3} and the overall loss function of the proposed method $\mathcal{L}_{KGAD}$ via Equation.~\ref{eq4}. By iteratively exploiting the gradient of $\mathcal{L}_{KGAD}$ to update the protective adversarial noise $\delta_i^{t}$ and clipping it to satisfy $\|\delta_i^{t}\|_{\infty} \leq \epsilon$, we can obtain the learned protective adversarial noise $\delta_i$ and corresponding adversarial sample $x_i^{\prime}$.

\begin{algorithm}[t]
\begin{small}
   \caption{Knowledge-Guided Adversarial Defense}
   \label{alg1}
\begin{algorithmic}[1]
   \REQUIRE Manipulation model $G_{\theta}$, number of dataset $N$, iteration number $T$ and perturbation budget $\epsilon$;
   \FOR{$i=1$ to $N$}
   \STATE Initialize protective adversarial noise $\delta_{i}^{0}$;
   \FOR{$t=0$ to $T-1$}
   \STATE Embed protective adversarial noise $\delta_i^{t}$ into $x_i$ and obtain the adversarial sample $x_i^{\prime_t}=x_i+\delta_i^{t}$;
   \STATE Forward-pass $x_i$ through $G_{\theta}$ and obtain the original fake sample $y_i=G_\delta(x_i)$;
   \STATE Forward-pass $x_i^{\prime_t}$ through $G_{\theta}$ and obtain the distorted fake sample $y_i^{\prime_t}=G_\delta(x_i+\delta_i^{t})$;
   \STATE Compute the loss function from domain-specific knowledge $\mathcal{L}_{dk}$ via Equation.~\ref{eq2} and the loss function from visual-perception knowledge $\mathcal{L}_{pk}$ via Equation.~\ref{eq3};
   \STATE Compute the overall loss function of the proposed method $\mathcal{L}_{KGAD}$ via Equation.~\ref{eq4};
   \STATE Compute the gradient of $\mathcal{L}_{KGAD}$ and update the protective adversarial noise $\delta_i^{t}$;
   \STATE Clip $\delta_i^{t}$ to satisfy $\|\delta_i^{t}\|_{\infty} \leq \epsilon$;
   \ENDFOR
   \STATE Output the learned protective adversarial noise $\delta_i$ and corresponding adversarial sample $x_i^{\prime}$;
   \ENDFOR
\end{algorithmic}
\end{small}
\end{algorithm}

\section{Experiment}
\label{sec4}
Empirical evaluations are conducted in this section. We first describe the experimental settings including datasets, manipulation models and hyperparameters. Afterwards, we report and analyze qualitative and quantitative results. Finally, we present ablation studies of the proposed method.

\subsection{Experimental settings}
\label{sec4.1}
\noindent\textbf{Datasets and manipulation models.}
We mainly utilize two datasets in our experiments: CelebA \cite{liu2015deep} and Monet2Photo \cite{zhu2017unpaired}. Since our goal is to protect the data during the inference phase, we only utilize their test data. In this work, we utilize StarGAN \cite{choi2018stargan}, AGGAN \cite{tang2019attention} and HiSD \cite{li2021image} as malicious face manipulation models. StarGAN is the most common visual editing model which are utilized by many defenses \cite{ruiz2020disrupting,aneja2022tafim} for validation. It is trained by using five attributes (black hair, blond hair, brown hair, gender and age) on CelebA according to the settings in \cite{ruiz2020disrupting}. AGGAN introduces the attention mechanism into face manipulation. It is trained with the same five attributes as StarGAN. HiSD is a latest face modification model, which is also trained on CelebA and can add a pair of glasses to the person or turn the hair black. We follow the settings in \cite{ruiz2020disrupting} and \cite{huang2021initiative} to evaluate the defenses. For StarGAN and AGGAN, we randomly select 50 face images from CelebA for testing defenses, \textit{i.e.}, 250 manipulations for each defense (because the manipulation model has five types of attribute modifications). For HiSD, we randomly select 250 face images from CelebA as the test data. In addition, for the malicious style manipulation, we use CycleGAN \cite{chu2017cyclegan} and AdaAttN \cite{liu2021adaattn} models as the manipulation models. CycleGAN is a common method to realize style transfer for images and AdaAttN is an advanced style manipulation model with an attention manner. We utilize their officially provided pre-training models (one type of style manipulation for CycleGAN and five types of style manipulations for AdaAttN) to evaluate the defenses on 50 images from the B test set of Monet2Photo.

\noindent\textbf{Evaluation indicators and baselines.}
 To evaluate the effectiveness of the defense methods, we utilize SSIM Dissimilarity (SSIMD), Feature Similarity Index Mersure Dissimilarity (FSIMD) and Learned Perceptual Image Patch Similarity (LPIPS) as the basic indicators. They are more consistent with human visual perception than the $L_2$-norm measure. 
 The work in \cite{hou2022perceptual} shows that substantial judgments via LPIPS are consistent with the human judgments when the LPIPS difference is greater than 0.40. Based on this observation, we set LPIPS $\geq$ 0.4 as a basic criterion for successfully defending against malicious manipulation models. This indicator is called as Success Rate (SR). To more clearly reflect the SR gap between defenses, we empirically modify the LPIPS threshold for different manipulation models (see specific experimental results). In addition, in order to measure the structural confusion of fake samples, we utilize an Canny-based edge detection technology \cite{canny1986computational,zhou2011improved} to obtain the contours of the original fake samples and distorted fake samples, and then calculate their $L_2$-norm distance. We call this indicator as $L_2^{con}$. Besides, for the malicious face manipulation, we perform face detection on the generated fake samples to evaluate whether their subsequent application or dissemination are effectively blocked. We exploit MTCNN \cite{zhang2016joint} as the detection model, and calculate the number of test samples whose faces are correctly detected. This domain-specific indicator is called as Blocking Rate (BR). Moreover, we leverage the iFLYTEK Starfire Cognitive Model as a cognitive model to understand the content of fake images to assess the extent to which the defenses perturb the fake samples and to check whether critical information (\textit{e.g.}, identity privacy) remains in the fake samples.

\begin{figure}[t]
\begin{center}
   \includegraphics[width=0.9\linewidth]{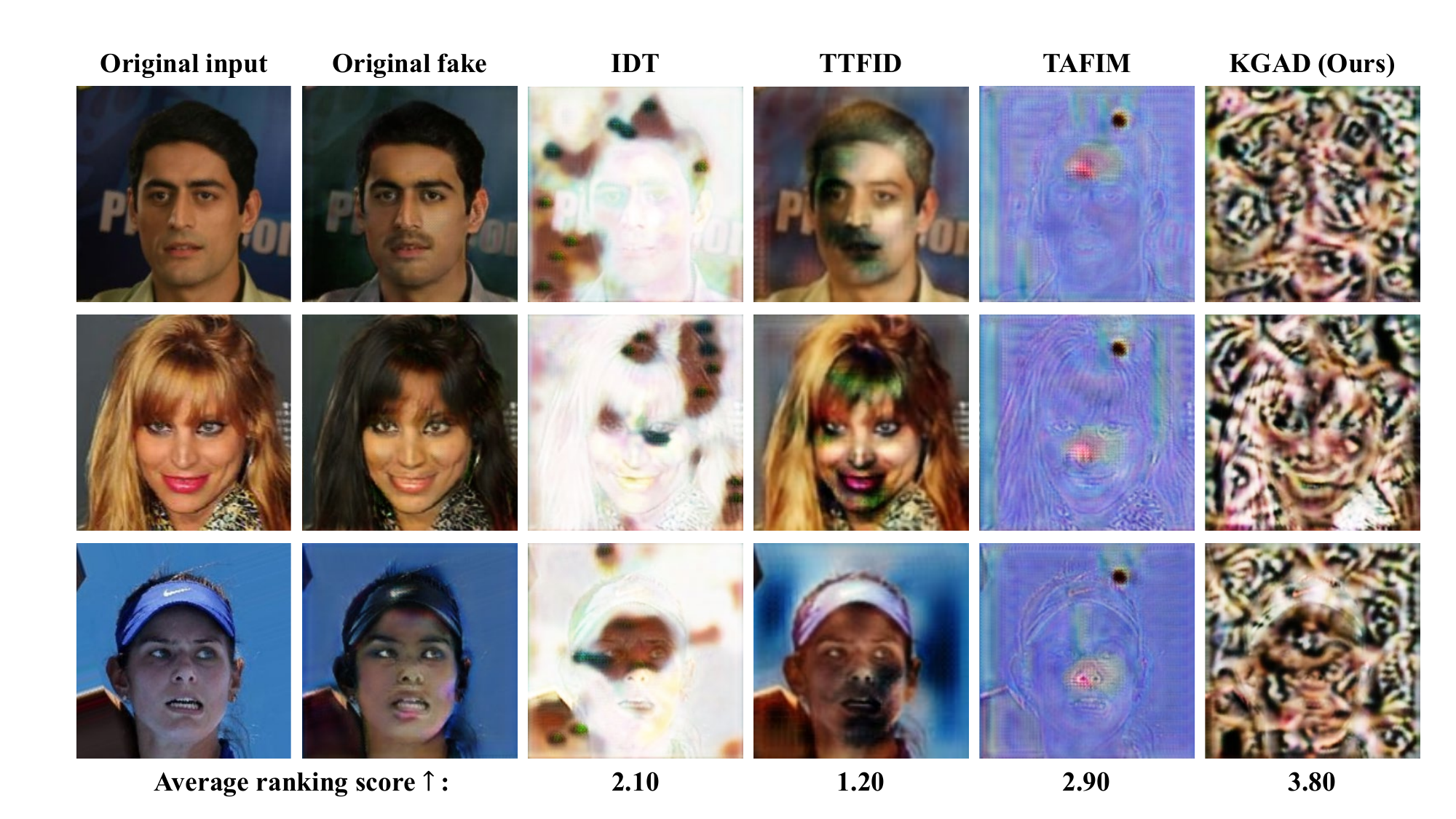}
\end{center}
   \caption{Examples of fake samples corrupted by different defense methods. The images in the first column are the original input samples, and the images in the right five columns are fake samples produced by the malicious manipulation model StarGAN. We utilize three defense methods as baselines: ITD \cite{ruiz2020disrupting}, TTFID \cite{huang2021initiative} and TAFIM \cite{aneja2022tafim}. We can observe that the perturbations caused by the baselines are mainly clustered in the color textures, and the face contours under the abnormal textures are still relatively clear. The face semantics of fake samples corrupted by our method are significantly perturbed, and the structures of the five senses become very confusing. In addition, we conduct a questionnaire to obtain feedback on image distortion from a perspective of human vision (the ranking score ranges from 1 to 4, a higher score indicates a stronger degree of the distortion).}
\label{fig5}
\end{figure}

\begin{figure}[t]
\begin{center}
   \includegraphics[width=0.9\linewidth]{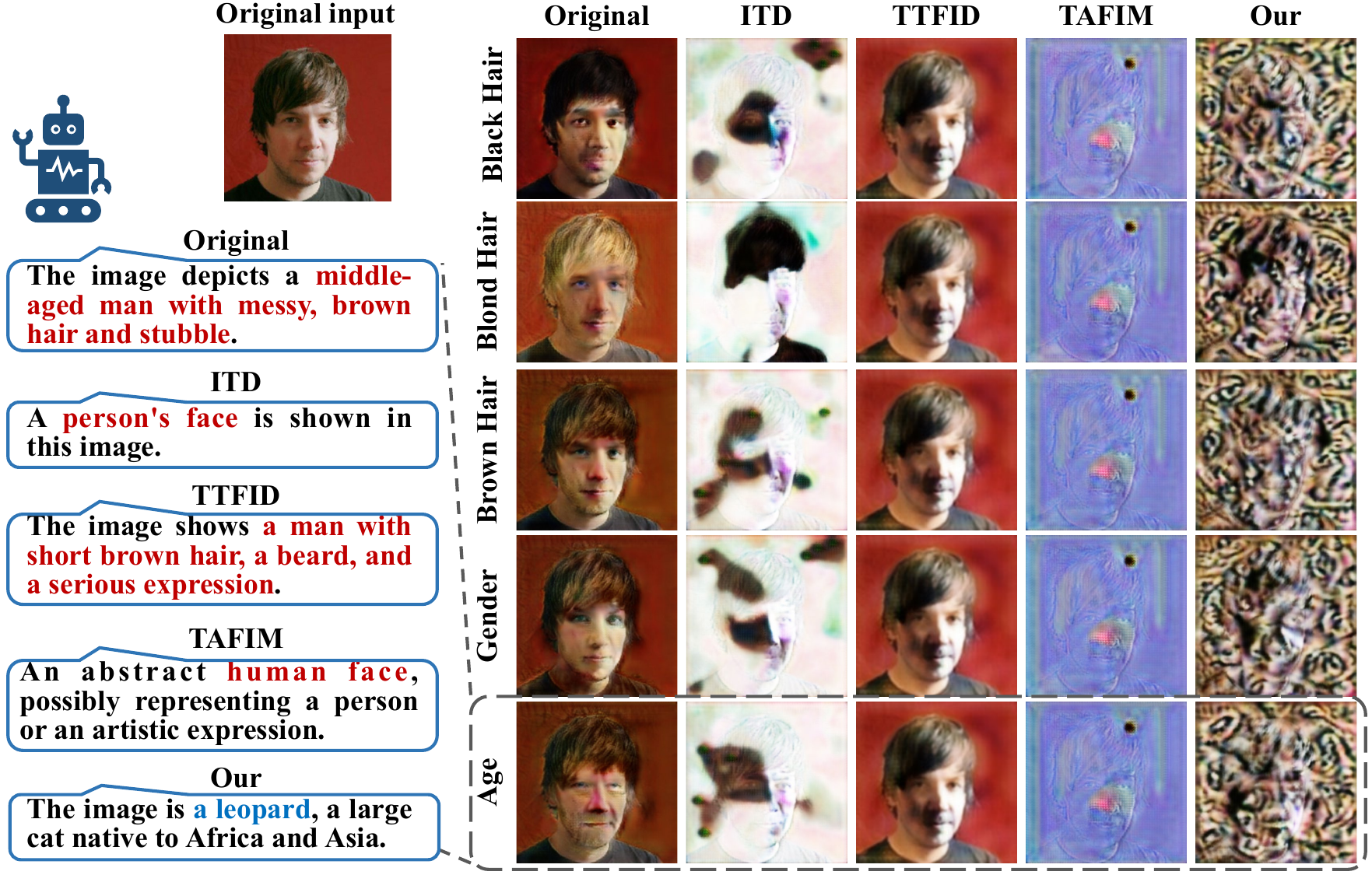}
\end{center}
   \caption{Examples of fake samples corrupted by different defense methods against StarGAN. The image in the upper left corner is the original input sample, and the images in the right five columns are fake samples produced by the manipulation model StarGAN. `Original' denotes the original fake sample. We use three defense methods as baselines: ITD, TTFID and TAFIM. We can observe that the perturbations caused by the baselines are mainly clustered in the color textures, and the face contours under the abnormal textures are still relatively clear. The face semantics of fake samples distorted by our method are significantly corrupted, and the structure of the five senses became very confusing. In addition, we also utilize a cognitive model to understand the content in the images (see the texts on the left). We take an overall attribute "age" as an example in this figure (the gray dotted box). The results show that this model does not recognize face-related content from the fake sample distorted by our method, which indicates that our method is effective in disrupting critical information and is thus able to protect identity privacy.}
\label{fig_s1}
\end{figure}

\begin{figure}[t]
\begin{center}
   \includegraphics[width=0.8\linewidth]{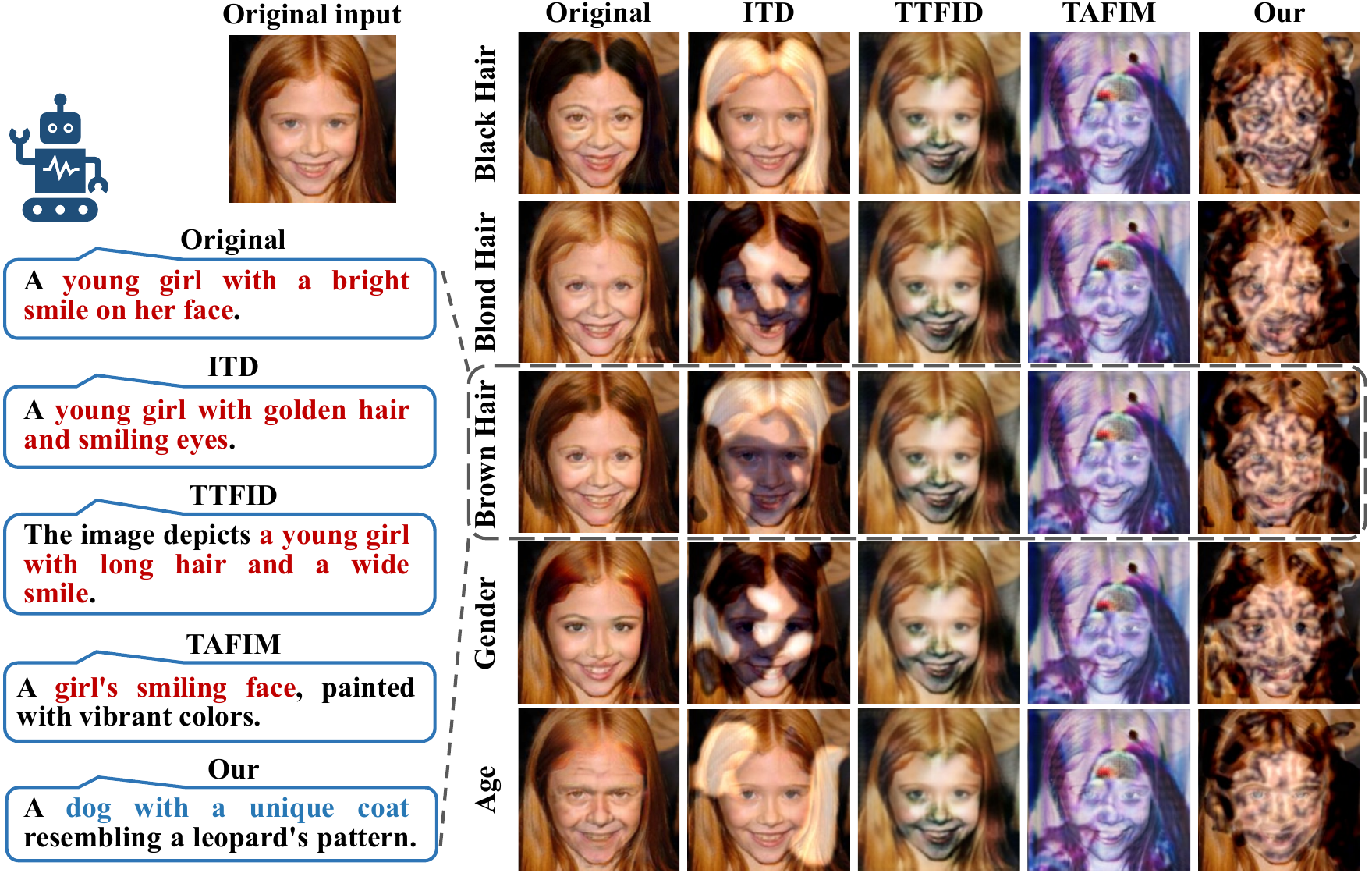}
\end{center}
   \caption{Examples of fake samples corrupted by different defense methods against AGGAN. The image in the upper left corner is the original input sample, and the images in the right five columns are fake samples produced by the manipulation model AGGAN. `Original' denotes the original fake sample. Similarly, we can observe that the perturbations caused by the baselines are mainly clustered in the color textures. In addition, we take the fake samples related to a local attribute "brown hair" as examples (the gray dotted box) for image understanding. The results show that the face-related content (\textit{e.g.}, "a girl") is not recognized from the fake sample distorted by our method.}
\label{fig_s2}
\end{figure}

\begin{figure}[b]
\begin{center}
   \includegraphics[width=0.8\linewidth]{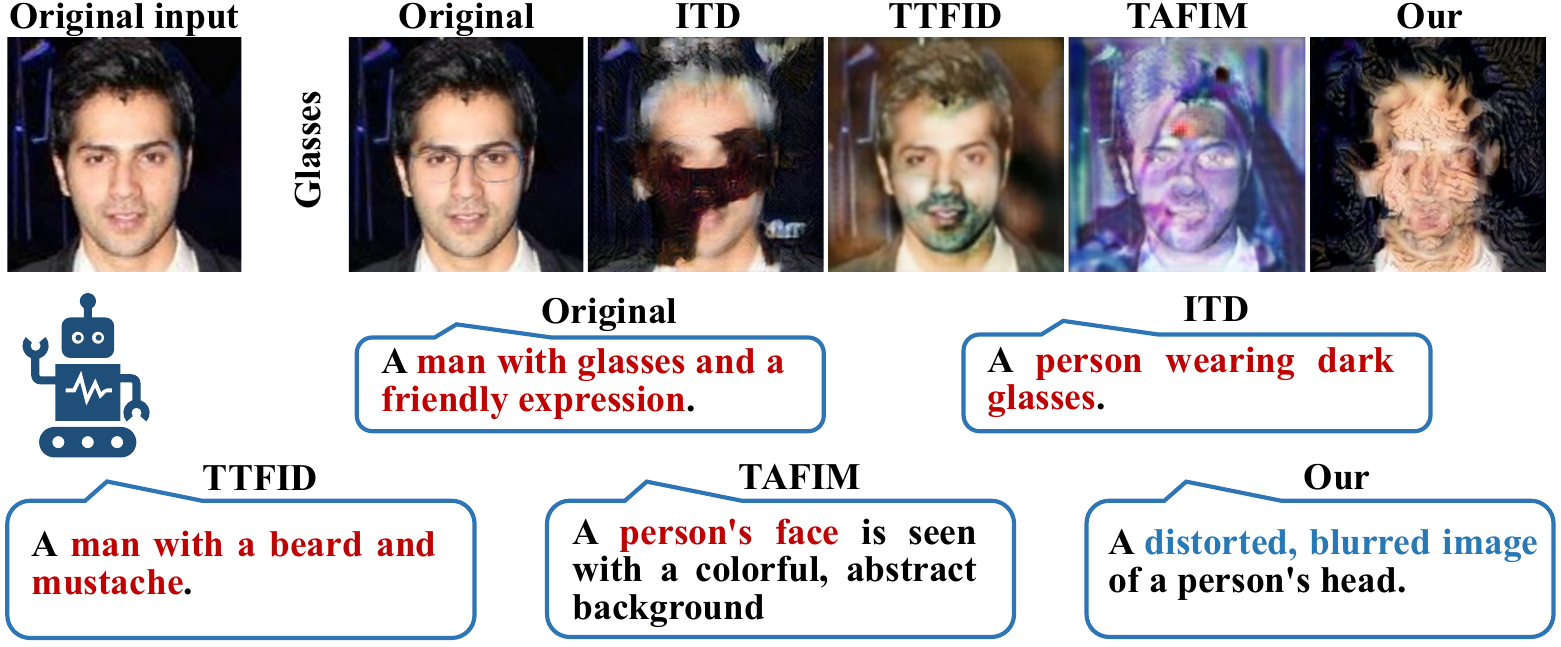}
\end{center}
   \caption{Examples of fake samples corrupted by different defense methods against HiSD. The image in the first column is the original input sample, and the images in the right five columns are fake samples produced by the manipulation model HiSD. `Original' denotes the original fake sample. The content understood by the intelligent model is shown below the images. It can be seen that the model recognizes information related to glasses and face from other fake samples, whereas confusing information is recognized from the fake sample corrupted by our method.}
\label{fig_s3}
\end{figure}

We take four representative defense methods as baselines: Image Translation Disruption (ITD) \cite{ruiz2020disrupting}, Two-stage Training Framework-based Initiative Defense (TTFID) \cite{huang2021initiative} and Targeted Adversarial attacks against Facial Image Manipulations (TAFIM) \cite{aneja2022tafim}. These methods cover different types of mechanisms for generating adversarial noise (\textit{e.g.}, gradient-based or generation model-based mechanisms) and have achieved advanced defensive effectiveness. The training settings of these defenses follow those in their original papers. Since the original papers of TTFID and TAFIM only focus on face manipulation, we just use ITD as the baseline against style manipulation. Note that all baselines For ITD and our defense, the step number is set to 60 and the step size is set to $1/255$. The perturbation budget is set to $7/255$ against the malicious face manipulation to make the size of their adversarial noise is similar to that generated by TTFID and TAFIM on MSE. The perturbation budget is $12/255$ against the malicious style manipulation. The hyperparameter $\lambda$ in our method is set to $1.0$ against the face manipulation and $6 \times 10^{-2}$ against the style manipulation.

\begin{figure}[t]
\begin{center}
   \includegraphics[width=0.65\linewidth]{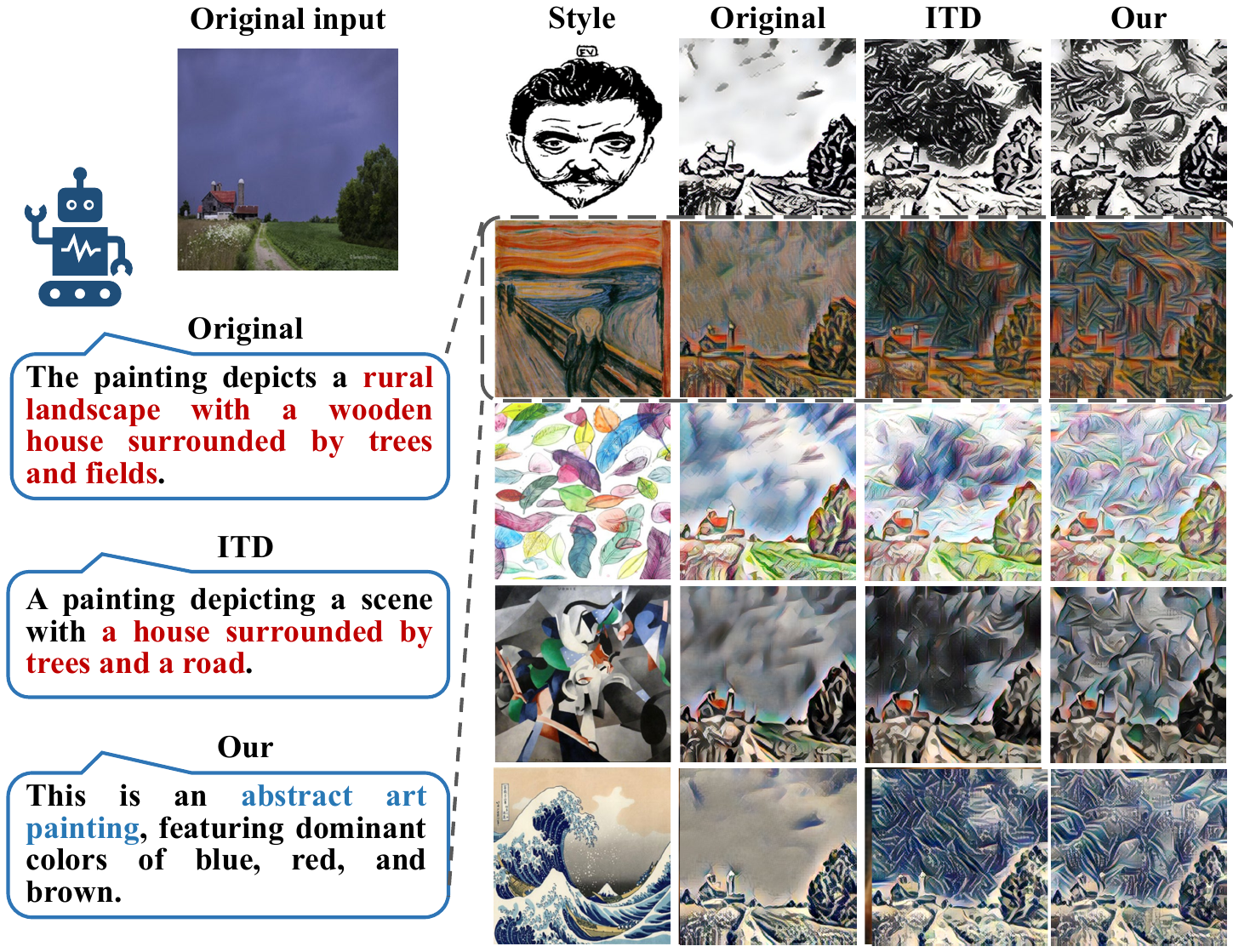}
\end{center}
   \caption{Examples of fake samples corrupted by different defense methods against AdaAttN. The image in the upper left corner (the first column) is the original input sample, the images in the second column are the target style samples and the images in the right three columns are fake samples produced by the manipulation model AdaAttN. `Original' denotes the original fake sample. We utilize ITD as the baseline. Compared with the baseline, our method produces greater damage to the semantic content (\textit{e.g.}, more obvious sharp mountain shapes in the sky). In addition, we exploit the cognitive model to understand the images. The results reflect that our method more fully obfuscates critical information (\textit{e.g.}, the house, road and trees).}
\label{fig_s4}
\end{figure}

\begin{figure}[b]
\begin{center}
   \includegraphics[width=0.55\linewidth]{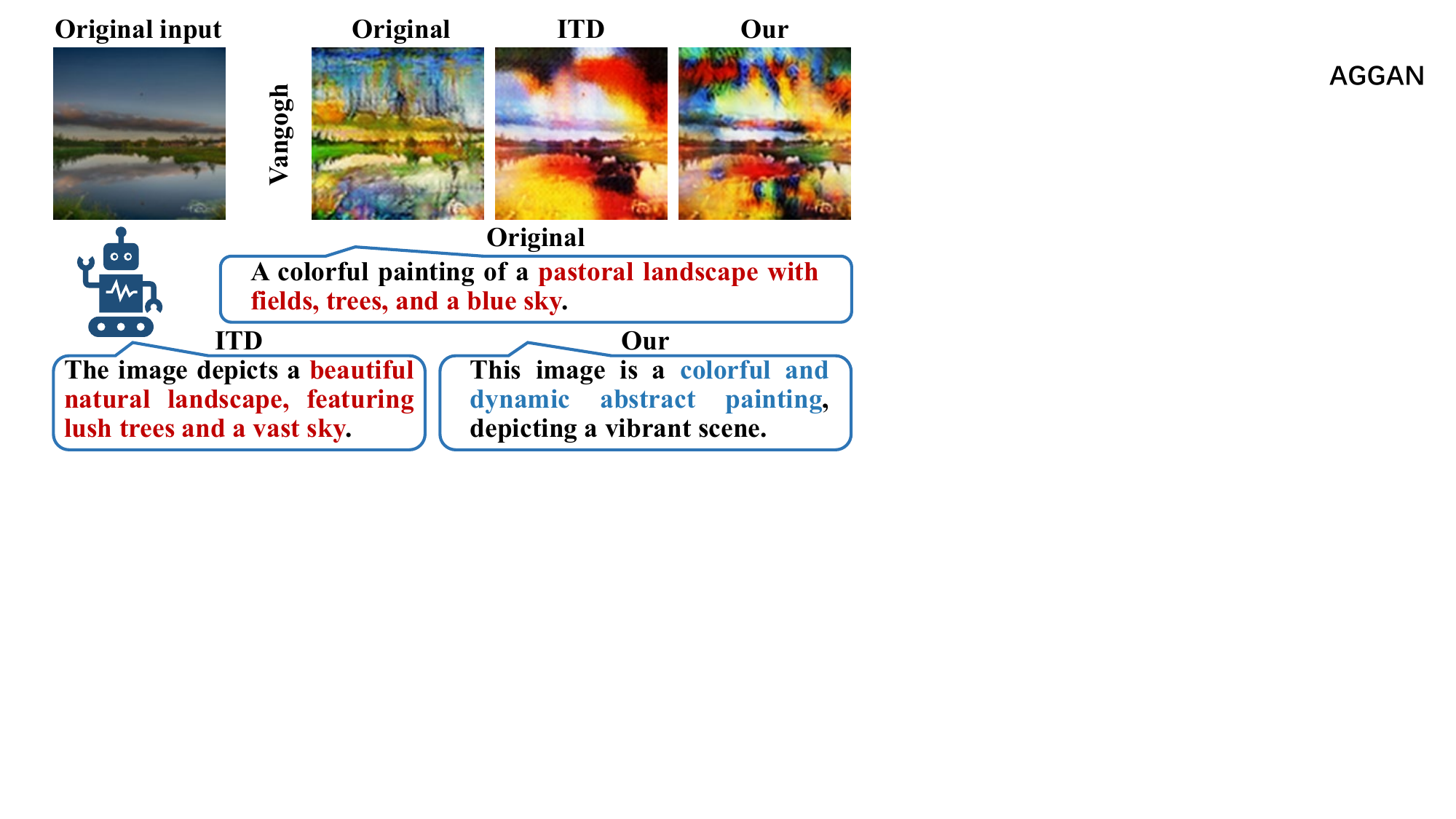}
\end{center}
   \caption{Examples of fake samples corrupted by different defense method against CycleGAN. The image from left to right are the original input sample and fake samples produced by the manipulation model CycleGAN. The target style is `Vangogh'. `Original' denotes the original fake sample. The content understood by the intelligent model is shown below the images. The information related to the trees and sky is not recognized in our method.}
\label{fig_s5}
\end{figure}

\subsection{Qualitative evaluation}
\label{sec4.2}
We first perform qualitative evaluations of defenses against both malicious face manipulation and malicious style manipulation. Taking the malicious face manipulation as an example, Figure~\ref{fig5} shows the distorted fake samples generated by StarGAN. We find that the perturbations caused by the baselines are mainly clustered in the color textures, and the face contours under the abnormal textures are still relatively clear. 
The face semantics of the fake samples corrupted by our defense are significantly perturbed, and the structures of the five senses become very confusing. In addition, we conduct a questionnaire to obtain feedback on image distortion from a perspective of human vision. The average ranking scores for each defense are shown at the bottom of the figure and the higher score indicates the the stronger degree of the distortion in samples. Our method gets the best score and has a large gap with other defenses (\textit{e.g.} 31\% improvement compared to the sub-optimal ranking score). 

In addition, more evaluations against different attribute manipulations are also conducted. Figure~\ref{fig_s1},~\ref{fig_s2},~\ref{fig_s3} show the fake samples with modified attributes generated by the face manipulation models StarGAN, AGGAN and HiSD, respectively. The images in the upper left corner (\textit{i.e.}, the first column) are the original input samples, and the images in the right five columns are fake samples produced by the malicious manipulation models. `Original' denotes the original fake sample. We can observe that the perturbations caused by the baselines are mainly clustered in the color textures, and the face contours under the abnormal textures are still relatively clear. The face semantics of fake samples distorted by our method are significantly corrupted, and the structures of the five senses also become very confusing. Moreover, we also leverage the cognitive model (iFLYTEK Starfire Cognitive Model) to understand the content in the images (see the texts in the figures). The results show that the cognitive model does not recognize face-related content from the fake sample corrupted by our method, which indicates that our method is effective in disrupting critical information and is thus able to protect identity privacy.

Figure~\ref{fig_s4} and Figure \ref{fig_s5} present the distorted fake samples generated by style manipulation models AdaAttN and CycleGAN. These results can demonstrate that our method can more significantly destroy the visual semantic content (\textit{e.g.}, more obvious sharp mountain shapes in the sky in Figure~\ref{fig_s4} and messy rips in the lake in Figure~\ref{fig_s5}), so that fake samples cannot be smoothly used for subsequent malicious actions. Inadequately, although our method forces the malicious manipulation model to produce more anomalous and confusing textures on this style manipulation task, the ability to obfuscate objects in the image does not yet reach the performance on the face manipulation task. This may be due to the fact that the used content extractor has not been powerful enough to accurately and comprehensively capture the shape semantics of the main object, making the generated adversarial noise not sufficiently disruptive to the object shape. Fortunately, the results of image understanding show that our method disrupts the important semantics in the fake samples to some extent, so that the critical information (\textit{e.g.}, the house and trees) in the fake samples cannot be distinguished.


\begin{table*}[t]
\begin{center}
\caption{The effects of defenses against different malicious face manipulation models (higher is better). The threshold of SR is set to 0.7, 0.5 and 0.5 for StarGAN, AGGAN and HiSD, respectively.}
\label{tab1}
\renewcommand\tabcolsep{7.0pt}
\renewcommand\arraystretch{1.2}
\begin{tabular}{cccccccc}
\hline
Manipulation Model & Defense & SSIMD $\uparrow$ & FSIMD $\uparrow$ & LPIPS $\uparrow$ & $L_2^{con}$ $\uparrow$ & SR $\uparrow$ & BR $\uparrow$ \\ \hline
\multicolumn{1}{c|}{} &ITD (\cite{ruiz2020disrupting}) & 0.4065 & 0.6326 & 0.6790 & 0.0842 & 30.80\% & 97.60\% \\
\multicolumn{1}{c|}{} &TTFID (\cite{huang2021initiative}) &  0.2001 & 0.2672 & 0.3522 & 0.1258 & 0.40\% & 30.00\% \\
\multicolumn{1}{c|}{} &TAFIM (\cite{aneja2022tafim}) & 0.1963 & 0.5864 & 0.7282 & 0.0716 & 67.20\% & 98.00\%  \\
\multicolumn{1}{c|}{\multirow{-4}{*}{StarGAN \cite{choi2018stargan}}} & \cellcolor[HTML]{C0C0C0} KGAD (Ours) & \cellcolor[HTML]{C0C0C0}\textbf{0.5276} & \cellcolor[HTML]{C0C0C0}\textbf{0.7013} & \cellcolor[HTML]{C0C0C0}\textbf{0.7354} & \cellcolor[HTML]{C0C0C0}\textbf{0.2027} & \cellcolor[HTML]{C0C0C0}\textbf{91.60\%} & \cellcolor[HTML]{C0C0C0}\textbf{100\%} \\ \hline
 \multicolumn{1}{c|}{}& ITD (\cite{ruiz2020disrupting}) & 0.2441 & 0.4135 & 0.3851 & 0.0629 & 6.80\% & 24.00\%   \\
\multicolumn{1}{c|}{} & TTFID (\cite{huang2021initiative}) & 0.1543 & 0.1724 & 0.2407 & 0.0821  & 0.00\% & 6.80\%  \\
 \multicolumn{1}{c|}{}& TAFIM (\cite{aneja2022tafim})  & 0.1704  &0.3915 &0.3713 &0.0530  &6.40\%  &23.20\%  \\
\multicolumn{1}{c|}{\multirow{-4}{*}{AGGAN \cite{tang2019attention}}} & \cellcolor[HTML]{C0C0C0} KGAD (Ours) & \cellcolor[HTML]{C0C0C0}\textbf{0.4088} & \cellcolor[HTML]{C0C0C0}\textbf{0.4501} & \cellcolor[HTML]{C0C0C0}\textbf{0.5165} & \cellcolor[HTML]{C0C0C0}\textbf{0.1116} & \cellcolor[HTML]{C0C0C0}\textbf{62.40\%} & \cellcolor[HTML]{C0C0C0}\textbf{89.60\%} \\ \hline
\multicolumn{1}{c|}{} & ITD (\cite{ruiz2020disrupting}) & 0.2962 & 0.5906 & 0.5268 & 0.1407 & 87.20\% & 78.80\% \\
 \multicolumn{1}{c|}{}& TTFID (\cite{huang2021initiative}) & 0.1527 & 0.2630 & 0.2583 & 0.1639  & 0.00\% &25.20\%  \\
\multicolumn{1}{c|}{} & TAFIM (\cite{aneja2022tafim}) &0.1634  &0.5176 &0.5312 &0.1290  &88.40\%  &80.40\%  \\
\multicolumn{1}{c|}{\multirow{-4}{*}{HiSD \cite{li2021image}}} & \cellcolor[HTML]{C0C0C0} KGAD (Ours) & \cellcolor[HTML]{C0C0C0}\textbf{0.4608} & \cellcolor[HTML]{C0C0C0}\textbf{0.6286} & \cellcolor[HTML]{C0C0C0}\textbf{0.5704} & \cellcolor[HTML]{C0C0C0}\textbf{0.2607} & \cellcolor[HTML]{C0C0C0}\textbf{97.20\%} & \cellcolor[HTML]{C0C0C0}\textbf{91.60\%} \\ \hline
\end{tabular}
\end{center}
\end{table*}

\begin{table*}[t]
\caption{The effects of defenses against different malicious style manipulation models (higher is better). The threshold of SR is set to $0.7$ and $0.4$ for CycleGAN and AdaAttN, respectively.}
\label{tab2}
\renewcommand\tabcolsep{10pt}
\renewcommand\arraystretch{1.2}
\begin{center}
\begin{tabular}{cccccccc}
\hline
Manipulation Model & Defense & SSIMD $\uparrow$ & FSIMD $\uparrow$ & LPIPS $\uparrow$ & $L_2^{con}$ $\uparrow$& SR $\uparrow$ \\ \hline
\multicolumn{1}{c|}{} & ITD (\cite{ruiz2020disrupting}) & 0.3018 & 0.4757 & 0.6376 & 0.4140 & 24.00\%   \\
\multicolumn{1}{c|}{\multirow{-2}{*}{CycleGAN \cite{zhu2017unpaired}}} & \cellcolor[HTML]{C0C0C0} KGAD (Ours) & \cellcolor[HTML]{C0C0C0}\textbf{0.3868} & \cellcolor[HTML]{C0C0C0}\textbf{0.4927} & \cellcolor[HTML]{C0C0C0}\textbf{0.6507} & \cellcolor[HTML]{C0C0C0}\textbf{0.4269} & \cellcolor[HTML]{C0C0C0}\textbf{26.00\%}  \\ \hline
 \multicolumn{1}{c|}{}& ITD (\cite{ruiz2020disrupting})& 0.2370 & 0.4288 & 0.4030 & 0.2869 & 49.60\%   \\
\multicolumn{1}{c|}{\multirow{-2}{*}{AdaAttN \cite{liu2021adaattn}}} & \cellcolor[HTML]{C0C0C0} KGAD (Ours) & \cellcolor[HTML]{C0C0C0}\textbf{0.2667} & \cellcolor[HTML]{C0C0C0}\textbf{0.4364} & \cellcolor[HTML]{C0C0C0}\textbf{0.4249} & \cellcolor[HTML]{C0C0C0}\textbf{0.2883} & \cellcolor[HTML]{C0C0C0}\textbf{62.80\%} \\ \hline
\end{tabular}
\end{center}
\end{table*}

\begin{table*}[t]
\caption{The defense against adversarially trained malicious manipulation models. The face manipulation model is StarGAN and the style manipulation model is CycleGAN. The threshold of SR is set to $0.3$. Since no face detection is performed in the style manipulation task, we do not report the value of BR for CycleGAN.}
\label{tab_s1}
\renewcommand\tabcolsep{7pt}
\renewcommand\arraystretch{1.2}
\begin{center}
\begin{tabular}{cccccccc}
\hline
Manipulation Model & Defense & SSIMD $\uparrow$ & FSIMD $\uparrow$ & LPIPS $\uparrow$ & $L_2^{con}$ $\uparrow$ & SR $\uparrow$ & BR $\uparrow$\\ \hline
 \multicolumn{1}{c|}{}& ITD & 0.1759 & 0.2467 & 0.2634 & 0.0317 &27.60\% &32.80\%  \\
 \multicolumn{1}{c|}{}& TTFID & 0.1160 & 0.1247 & 0.1865 & 0.0612 &0.00\% &9.20\%  \\
 \multicolumn{1}{c|}{}& TAFIM & 0.1039 & 0.2147 & 0.3087 & 0.0298 &57.60\% &33.20\%   \\
\multicolumn{1}{c|}{\multirow{-4}{*}{StarGAN}} & \cellcolor[HTML]{C0C0C0} KGAD (Ours)& \cellcolor[HTML]{C0C0C0}\textbf{0.3063} & \cellcolor[HTML]{C0C0C0}\textbf{0.3780} & \cellcolor[HTML]{C0C0C0}\textbf{0.4132} & \cellcolor[HTML]{C0C0C0}\textbf{0.1176} &\cellcolor[HTML]{C0C0C0}\textbf{67.20\%} &\cellcolor[HTML]{C0C0C0}\textbf{42.40\%} \\ \hline
 \multicolumn{1}{c|}{}& ITD & 0.1447 & 0.2319 &0.2967 &0.2018 &48.40\% & -\\
\multicolumn{1}{c|}{\multirow{-2}{*}{CycleGAN}} & \cellcolor[HTML]{C0C0C0} KGAD (Ours)& \cellcolor[HTML]{C0C0C0}\textbf{0.2362} & \cellcolor[HTML]{C0C0C0}\textbf{0.3130} & \cellcolor[HTML]{C0C0C0}\textbf{0.3506} & \cellcolor[HTML]{C0C0C0}\textbf{0.2279} &\cellcolor[HTML]{C0C0C0}\textbf{55.20\%} &\cellcolor[HTML]{C0C0C0}-
\\ \hline
\end{tabular}
\end{center}
\end{table*}

\subsection{Quantitative evaluation}
\label{sec4.3}
In addition to the above qualitative evaluations, we also quantitatively evaluate the effectiveness and generalization of the defense methods from the aspects of distortion, universality, and transferability.

\begin{table}[t]
\caption{The universality of adversarial noise generated by defenses against the face manipulation model StarGAN (higher is better). The threshold of SR is set to $0.6$.}
\label{tab3}
\renewcommand\tabcolsep{7pt}
\renewcommand\arraystretch{1.2}
\begin{center}
\begin{tabular}{cccccc}
\hline
Defense & SSIMD $\uparrow$ & LPIPS $\uparrow$ & $L_2^{con}$ $\uparrow$ & SR $\uparrow$ & BR $\uparrow$ \\ \hline
\multicolumn{1}{c|}{ITD}  & 0.2459  & 0.5976 &0.0736  & 46.40\% & 46.30\% \\
\multicolumn{1}{c|}{TTFID}  & 0.1728  & 0.2969 & 0.1034  & 0.00\% & 26.00\%  \\
\multicolumn{1}{c|}{TAFIM}  & 0.1853  & 0.6518 & 0.0702  & 76.00\% & 62.00\% \\
\rowcolor[HTML]{C0C0C0} 
\multicolumn{1}{c|}{KGAD (Ours)} & \textbf{0.3681} & \textbf{0.6606} & \textbf{0.1707} & \textbf{96.40\%} & \textbf{100\%} \\ \hline
\end{tabular}
\end{center}
\end{table}

\begin{table}[t]
\caption{The transferability of adversarial noise against different face manipulation models. The source model is StarGAN and the target models are AGGAN and HiSD.}
\label{tab4}
\renewcommand\tabcolsep{7pt}
\renewcommand\arraystretch{1.2}
\begin{center}
\begin{tabular}{cccccc}
\hline
Model & Defense & SSIMD $\uparrow$ & FSIMD $\uparrow$ & LPIPS $\uparrow$ & $L_2^{con}$ $\uparrow$ \\ \hline
 \multicolumn{1}{c|}{}& ITD & 0.4079 & 0.6374 & 0.6727 & 0.0728   \\
 \multicolumn{1}{c|}{}& TTFID & 0.2196 & 0.2851 & 0.3683 & 0.1201  \\
 \multicolumn{1}{c|}{}& TAFIM & 0.3175 & 0.6041 & 0.7360 & 0.0626   \\
\multicolumn{1}{c|}{\multirow{-4}{*}{StarGAN}} & \cellcolor[HTML]{C0C0C0} Ours& \cellcolor[HTML]{C0C0C0}\textbf{0.5229} & \cellcolor[HTML]{C0C0C0}\textbf{0.7037} & \cellcolor[HTML]{C0C0C0} \textbf{0.7372} & \cellcolor[HTML]{C0C0C0}\textbf{0.1928}  \\ \hline
 \multicolumn{1}{c|}{}& ITD & 0.1856 & 0.3721 &0.3376 &0.0529\\
 \multicolumn{1}{c|}{}& TTFID & 0.1247 & 0.1495 & 0.2270 & 0.0530 \\
 \multicolumn{1}{c|}{}& TAFIM & 0.1549  &0.3630 &0.3082 &0.0498      \\
\multicolumn{1}{c|}{\multirow{-4}{*}{AGGAN}} & \cellcolor[HTML]{C0C0C0} Ours& \cellcolor[HTML]{C0C0C0}\textbf{0.1903} & \cellcolor[HTML]{C0C0C0}\textbf{0.4217} & \cellcolor[HTML]{C0C0C0}\textbf{0.3700} & \cellcolor[HTML]{C0C0C0}\textbf{0.0615} \\ \hline
 \multicolumn{1}{c|}{}& ITD & 0.0264 & 0.1752 & 0.1323 & 0.0331 \\
 \multicolumn{1}{c|}{}& TTFID & 0.0104 & 0.0515 & 0.0261 & 0.0284 \\
 \multicolumn{1}{c|}{}& TAFIM &0.0170  &0.1421 &0.0768 &0.0278    \\
\multicolumn{1}{c|}{\multirow{-4}{*}{HiSD}} & \cellcolor[HTML]{C0C0C0} Ours & \cellcolor[HTML]{C0C0C0} \textbf{0.0280} & \cellcolor[HTML]{C0C0C0} \textbf{0.1783} & \cellcolor[HTML]{C0C0C0}\textbf{0.1590} & \cellcolor[HTML]{C0C0C0}\textbf{0.0367}\\ \hline
\end{tabular}
\end{center}
\end{table}

\noindent\textbf{Distortion Assessment.}
We generate the adversarial noise against four malicious face manipulation models and two malicious style manipulation models. As shown in Table~\ref{tab1} and Table~\ref{tab2}, our method achieves the highest scores on several indicators, which indicates that our method had better protection  capability for visual data. In addition, we evaluate the defensive effect against adversarially trained manipulation models. These manipulation models are optimized by introducing additional adversarial data. We use a 10-step PGD with a perturbation of 8/255 and a step size of 2/255 to generate adversarial data for training the manipulation models. The results are shown in Table~\ref{tab_s1}. Although these results are not as high as those against normally trained malicious manipulation models, our method still achieves great and leading performances. Moreover, adversarial training may affect the normal performance of the manipulation model \cite{goodfellow2014explaining,madry2017towards}, which relatively limits the application of the manipulation model. Therefore, malicious models in real scenarios are usually not adversarially trained. Our method is expected to provide effective protection for visual data in the real world.



\noindent\textbf{Universality of adversarial noise.}
Generating specific adversarial noise for each sample usually requires a lot of time consumption, so protective adversarial noise is expected to be universal to different input samples. We apply the adversarial noise generated for one input sample to other 249 input samples. The results shown in Table~\ref{tab3} present that our method has competitive universality. Note that our method has no special design for the universality and it mainly exploits the guidance of the domain-specific knowledge and visual-perception knowledge.



\noindent\textbf{Transferability of adversarial noise.}
The transferability of protective adversarial noise is also an important characteristic that deserves attention (a type of capability in the black-box scenario). We randomly select 250 images from CelebA and then generative adversarial noise against StarGAN. To be consistent with HiSD, StarGAN only performs one type of face manipulation here (\textit{i.e.}, turning the hair black). We then feed these protected samples into other malicious manipulation models. The results in Table~\ref{tab4} show that our method still has competitive performance without modules specially designed for the transferability.



\begin{figure}[t]
\begin{center}
   \includegraphics[width=0.6\linewidth]{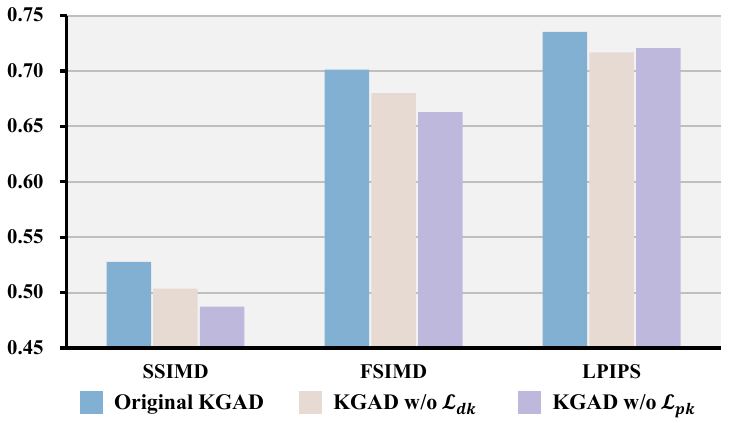}
\end{center}
   \caption{The illustration of the ablation study. We evaluate the effects of different terms in our method by removing $\mathcal{L}_{dk}$ and $\mathcal{L}_{pk}$, respectively. We find that introducing these terms has positive effects on protecting visual data.}
\label{fig6}
\end{figure}

\subsection{Ablation studies}
\label{sec4.4}
To evaluate the effects of different terms in the proposed method, we remove the loss function based on domain-specific knowledge (\textit{i.e.}, $\mathcal{L}_{dk}$) and the loss function based on visual-perception knowledge (\textit{i.e.}, $\mathcal{L}_{pk}$), respectively. As shown in Figure~\ref{fig6}, we find that both of them play positive roles in defending against malicious manipulation models. In addition, we note that the guidance from domain-specific knowledge has a greater boost to the improvement of LPIPS.


\section{Conclusion}
\label{sec5}
With the improvement of deep generation technologies, malicious applications of visual manipulation have raised serious security and reputation threats in the society. To mitigate this issue and protect visual data, a type of active defensive mechanism based on adversarial noise has received increasing attention. However, such defenses usually belong to ``data-only" methods and the important knowledge in visual manipulation has not been well exploited. 
Frontier researches have shown that integrating knowledge in deep learning can promote a focus on high-level semantics, yielding reliable and generalizable solutions. Inspired by this, we 
propose a knowledge-guided adversarial defense. By maximizing the disruptions at the level of domain-specific and visual-perception knowledge, the generated adversarial noise is expected to interfere with malicious manipulation models to produce significant semantic confusions in fake samples, thus preventing the commission of malicious actions. Qualitative and quantitative experiments demonstrate the effectiveness of the proposed defense, and the evaluation on the image understanding indicates that our method can more effectively obfuscate critical information (\textit{e.g.}, identity privacy) in fake samples to mitigate the damage caused by potentially malicious actions in the real world. 

\noindent Limitation: Most related works are currently based on white-box scenarios, similarly, our method has not yet covered a special design for the transferability or universality. Fortunately, relying on the own knowledge guidance, our method achieves better generalizability than baselines. In future work, we will introduce additional mechanisms (\textit{e.g.}, distribution manipulation \cite{zhu2022toward} and manifold attack model \cite{chen2024theory}) to further improve the generalization and black-box capabilities. Overall, this work is dedicated to providing a new insight for the defense against the malicious deepfake to further protect visual data, and we hope to inspire more great works for this worthwhile but not yet sufficiently explored field.



%

\bibliographystyle{unsrt}  
\bibliography{reference}

\end{document}